# Machine Learning Framework: Competitive Intelligence and Key Drivers Identification of Market Share Trends Among Healthcare Facilities

Anudeep Appe, Bhanu Poluparthi, Lakshmi Kasivajjula, Udai Mv, Sobha Bagadi, Punya Modi, Aditya Singh, Hemanth Gunupudi

*Abstract*—The necessity of data driven decisions in healthcare strategy formulation is rapidly increasing. A reliable framework which helps identify factors impacting a Healthcare Provider Facility or a Hospital (from here on termed as Facility) Market Share is of key importance. This pilot study aims at developing a data driven Machine Learning - Regression framework which aids strategists in formulating key decisions to improve the Facility's Market Share which in turn impacts in improving the quality of healthcare services. The US (United States) healthcare business is chosen for the study; and the data spanning across 60 key Facilities in Washington State and about 3 years of historical data is considered. In the current analysis Market Share is termed as the ratio of facility's encounters to the total encounters among the group of potential competitor facilities. The current study proposes a novel two-pronged approach of competitor identification and regression approach to evaluate and predict market share, respectively. Leveraged model agnostic technique, SHAP, to quantify the relative importance of features impacting the market share. Typical techniques in literature, to quantify the degree of competitiveness among facilities, use an empirical method to calculate a competitive factor to interpret severity of competition. The proposed method to identify pool of competitors, develops Directed Acyclic Graphs (DAGs), feature level word vectors and evaluates the key connected components at facility level. This technique is robust since its data -driven which minimizes the bias from empirical techniques. The DAGs factor in partial correlations at various segregations, key demographics of facilities along with a placeholder to factor in various business rules (for ex. quantifying the patient exchanges, provider references, sister facilities ...). Post identifying the set of competitors among facilities, developed and fine-tuned Random Forest based Regression model to predict the Market share. To identify key drivers of market share at an overall level, permutation feature importance of the attributes was calculated. For relative quantification of features at a facility level, incorporated SHAP (SHapley Additive exPlanations) a model agnostic explainer. This helped to identify and rank the attributes at each facility which impacts the market share. This approach proposes a unique amalgamation of the two popular and efficient modeling practices viz., Machine Learning with graphs and tree-based regression techniques to reduce the bias. With these, we helped to drive strategic business decisions.

## I. Introduction

The healthcare industry is ever changing, evolving, and progressing at a rate unlike any other industry. Given this dynamic nature, it has become more important for healthcare strategists to come up with plans that account for the unknowns and knowns in the system while understanding their impact.
A hospital strategic planning would involve like any other strategic planning, goals, and objectives and thus construction of a plan to achieve them. While external factors like government policies, technological advancements and economic trends play a crucial role in setting the goals and achieving them for a typical hospital facility, other significant impacting factors are competitive intelligence and market share impactors.

The analysis aims to assist the healthcare facilities with competitive intelligence, an effective strategy to understand the potential factors which are driving the target i.e., Market Share. The goal of good competitive strategy is not to enable a firm to mimic the strategy of its competitors; instead, it should be used to anticipate competitor actions and seek ways to achieve or maintain superior competitive positioning. In constructing such a plan, typical challenges that strategists would face are not knowing the right competitors, unable to understand what drives the encounters (Encounters are defined as visits involving interaction of patient and a provider who exercises independent judgment in the provision of services to the individual) or market share of a facility (Ratio of facility encounters divided by total encounters in that state) and no visibility into future scenarios and their impact. In this paper we tried to address the above-mentioned challenges of competitor identification and drivers of market share.

After a thorough literature review and consulting domain experts, decided to tackle this problem using a two-pronged approach. Leveraging the existing studies and formulating the data driven techniques, the paper proposes a two-step data driven approach to identify the competitors among the set of facilities and then to identify the key drivers impacting the Market Share.

## II. Related Work

There are multiple studies for analyzing health care market size and to quantify degree of competition for healthcare facilities[1]. Various fronts include segregation of the required facilities on the geographical locality, population density, healthcare services provided by the facilities [2] ... Observed that majority of the existing approaches address the competition among the facilities using varied empirical equations with primary emphasis on distance between two facilities. Widely accepted and used measure of competition in literature is Hirschman-Herfindahl index (HHI) is based on empirical study[3] The current study has taken the parallels from factors used in the HHI index study and developed a data driven Graph based approach for identifying the pool of competitors[4].

Having the appropriate techniques to have the explainable and understandable predictions made by complex Machine Learning models is of paramount importance. This field of study is formally termed as Explainable AI (Artificial

All Authors are associated with Providence India LLC, Hyderabad, 500032 TS, India (e-mail: <firstname.lastname>@ providence.org).

Intelligence). Understanding the AI (Artificial Intelligence) / ML (Machine Learning) output helps to build trust and increases the actionability while handling real-world problems with improved decision-making strategy. Various techniques like permutation feature importance[5], tree interpreters LIME[6], SHAP[7] (SHapley Additive exPlanations) are widely used for analyzing the key explanatory features impacting the target variable. All these are model agnostic techniques and help in relative ranking of features' impact on the outcome. The current study uses a Regression based approach to predict the market share and then uses the trained model to identify the key drivers impacting the market share. The paper uses the SHAP technique and extends the SHapley values calculations to identify and interpret the effect of the key features on the Market Share at various segregations.

III. DATASET

Fig. 3.1 illustrates the flow of data. The dataset created to facilitate the functioning of this project covers the seven states where Providence Healthcare has a presence, spans a period of six years, stretching from 2016 to 2022 and contains over 45million entries, each of which corresponds to a patient encounter which has occurred at a healthcare facility present in one of the states we have sampled.

The dataset creation process began with identifying all the disparate, high-quality data sources which contained information relevant to the project's purpose. The data so identified after a thorough and rigorous data exploration undertaking is subsequently processed through the requisite extraction, transformation and loading processes which prepares the data for the data model.

Fig. 3.2 represents the Entity Relationship diagram of the data model. An entity relation diagram is a visual tool which helps individuals better understand the objects existing within a defined system and how they correspond, relate, and interact with one another. Due to data compliance and privacy reasons, all the attributes inhabiting the tables cannot be displayed in the above illustration. The advantage of modelling the dataset are as follows

1) Improves the documentation and standardization of various data sources.
2) This leads to improvements in query runtimes and space complexity.
3) Increases scalability and adaptability of the data set.

The data model utilized for this project was a snowflake schema with critical fact tables being referenced by multiple dimension tables which in turn link to other sub-dimension tables. Fact tables also termed as Reality tables which contain quantitative information in a de-normalized form while dimension tables contain the dimensions along which the values of the attributes are taken in the fact table. A snowflake schema ensures lesser data redundancy and consumption of space.

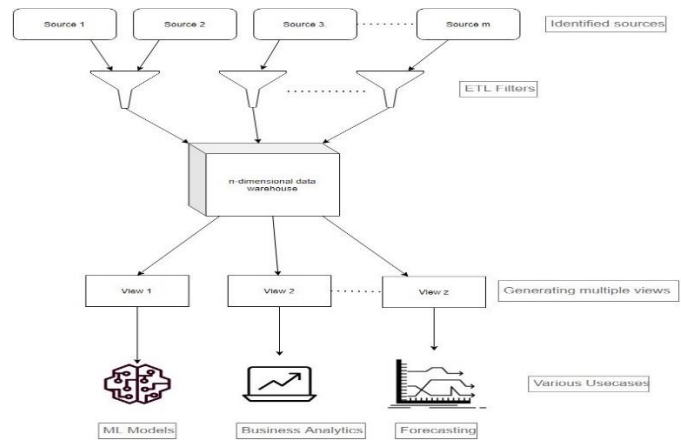

Fig. 3.1 Data Model Architecture

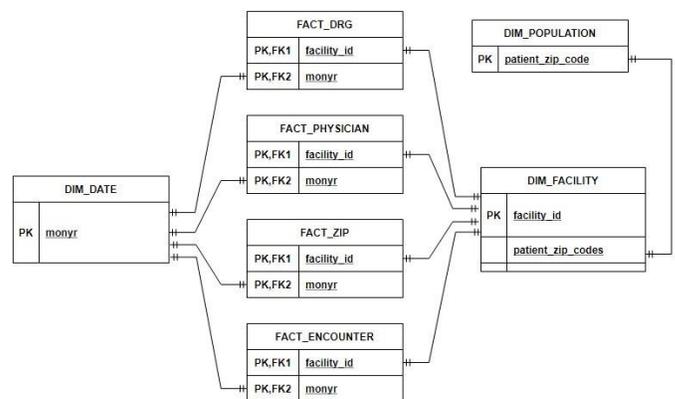

Fig. 3.2 Entity Relationship diagram

The tables used in the model are as follows

A. *Fact Tables:*

1) Encounter: Contains critical data pertaining to each encounter recorded at a facility regarding payor and hospital department where the encounter occurred. A composite key comprising of the facility id and month/year forms a primary key to access data in the table.

2) DRG: Contains data for multiple disease groupings such as their encounter numbers and rank for each facility at a monthly level.

3) ZIP: Contains data about the number of encounters logged from that zip code and the ranking of the zip codes per facility at a monthly frequency.

4) Physician: the number of encounters recorded by physicians, at a facility for a given month.

Multiple measures were taken to follow the guidelines of HIPAA (Health Insurance Portability and Accountability) compliance in handling PHI (Personal Health Information) data. The study uses de-identified, masked data without any PHI related information

*B. Dimension Tables:*

1) Facility: - Crucial metrics such as bed count, facility and aggregated reviews, hospital type, count of competing facilities, the facility's service area etc. are contained in this table.

2) Population: - It includes data on population growth projections, base population of that zip code and actual population growth recorded.

3) Date: - Contains the month, year, and month/year columns where each month corresponds to the month for which the data was harnessed.

*C. Feature Engineering:*

For all the approaches, it was determined that a set of auxiliary features was required to complement the pre-existing numerical features to improve the models, which were created using feature engineering [8]. This resulted in the creation of several supplementary features, a brief overview of the input features is as follows:

1) Encounters: are defined as an admission of a patient at a healthcare facility during which the Medical Staff Member has direct, in-person contact with the patient. Besides total encounters, several subsets of encounters have been included as input features as mentioned below

   a) CPS Institute-based encounters – Clinical Program Service incorporates departments and sub departments in the hospital that acts as a service line for the hospital business unit. Encounters were sorted based on CPS Institute with the dataset showing month wise encounters based on CPS Institute

   b) Age group-based encounters – Encounters were sorted based on age intervals

   c) Competitor-based encounters – Competitors were identified based on (discussed in section (V.A)) and their encounters were merged to get the month wise total number of encounters for competitor facilities

2) Hospital Attributes: A patient's choice of hospital depends upon several factors including physical attributes of the facility [9], internal analysis done by the business strategy team has shown several facility wise attributes that effect the encounters of a facility that are mentioned below

   a) Licensed Bed Count – Studies have shown Licensed hospital Beds can directly affect the number of patients encounters possible [12]

   b) Number of Nurses - Studies have shown that number of nurses can directly affect the number of patients treated and patient satisfaction [13]

   c) Service Area – A geographical grouping decided upon by the business strategy team of the Providence group of hospitals. While fitting the model we fed this Service Area feature doing one hot encoding

   d) Covid Facility – A Boolean that showed if the facility accepted patients diagnosed with Covid-19. Several studies have shown facilities treating covid19 encounters have seen differences in encounter trends as well as patients' experiences[11]

3) Rankings – To reduce the dimensionality of the dataset and to identify the top contributors to the number of encounters and market share of the facility, features were categorically grouped and the top n contributors to encounters, the choice of n was evaluated on similar lines of pareto analysis principle, to identify the 'n' contributors which have major effect on the metric chosen. Output was pivoted around facility and timeframe and added to the dataset. This process was done for the following features –

   a) Zip Codes – An integer value that showed the number of encounters that Originated from the specified Zip Code in a month

   b) Physician Encounters – An integer value that showed the number of encounters attended to by a particular physician in a month

   c) Disease Related Groupings Encounters – An integer value that showed the number of encounters for a particular disease grouping in a month

4) Patient Satisfaction and Experience – Patient satisfaction and experience plays a critical role [14] in retaining existing patients and gaining new patients.

   a) Sentiment Analysis of reviews – Research has shown negative reviews** can impact the encounters at a facility[15]. To combat this and improve patient experience, reviews consisting of a rating out of 5 stars as well as an optional written review were acquired. The review text was run through an inhouse models of parts of speech tagging and sentiment analysis that gave the text a sentiment score between 0 and 5, also tagged the phrase that contributed to the sentiment score and the subject the phrase was directed to. This was incorporated into the dataset by including the month wise no of positive and negative reviews and text.

   b) HCAHPS survey - HCAHPS[16] is a standardized, publicly reported survey of patients' perspectives of hospital care. It is a 29-item instrument and data collection methodology for measuring patients' perceptions of their hospital experience. This survey was incorporated into the dataset by adding nurse ratings and patient experience

** Appropriate measures were taken to maintain the anonymity of the reviews used in the analysis

## IV. PROBLEM STATEMENT

Each entry to the dataset is created at Market Share calculated at the facility and month level. Market Share is the ratio of encounters of a facility in a specified period to the ratio of total encounters of all the competitors considered in a particular connected component. For the analysis considered the state hospital data which has the encounter information shared by various facilities. The ratio is multiplied by 100 to convert to percentage. All the features created in feature engineering step are aggregated at the month level and are merged.

Now given a dataset Z with a one-to-one mapping of Market share(y) with all the explanatory features set (X) is created and the learning problem is formulated as the to predict y given X.

As the target y is continuous the learning problem falls under the regime of Regression approach. The choice of loss function to minimize the error considered is minimizing the mean squared error between actual & predicted values. This is the learning goal of the problem.

Depending on the nature of the algorithm chosen, the predicted values are capped between 2% and 99%. These extreme limits are chosen for countering the high outliers in prediction which potentially cause the higher magnitude of errors and impact the readability of predictions.

Considered 24 months (about 2 years) of data for training the model and used subsequent 3 months of data for testing the model. The train and test split cannot be implemented as random split as the records are at monthly intervals and will affect the data and model's continuity.

## V. SOLUTION FORMULATION

### A. Competitor Identification

Competitive research is an essential component of strong marketing strategies. There have been multiple attempts made towards identifying primary competitors in the market of various businesses. In the lens of healthcare providers, the market share is the distribution of encounters spread across various healthcare facilities. The basic understanding of sharing the market in a healthcare facility would be to have patients visit those facilities from the same areas, to avail similar services, with a similar set of payor channels, with similar or comparable healthcare plans etc... To structure the process of identifying competitors, an approach was devised using Directed Acyclic Graphs (DAGs) and feature level word vectors. This approach brings into consideration distinct factors that affect the competition between two facilities.

Competitor Identification Process: The approach uses partial correlation of the encounters at a facility level. It can be understood through Fig. 5.1.

The competitor identification process is as follows

1) Calculate the distances between all facilities: The distance between two facilities or a group of facilities can influence the market share of those facilities. The distances between all the facilities in the dataset were calculated using geopy.

2) Partial correlation between facilities at the service line: Statistically partial correlation measures the degree of association between two random variables with the effect of a set of controlling random variable[17]. Using pingouin, a python library partial correlation of monthly encounters at each facility for each service line is calculated.

3) Filtering the facilities based on distance measures and partial correlation:

   a) The distance measures and partial correlation was used to extract highly affecting facilities. The highly correlated facilities i.e., with the partial correlation values greater than 0.8 and less than -0.7 were filtered. Optimal cutoff values were chosen through thorough hyperparameter tuning.

   b) As per Subject Matter Experts (SME) review added a filter of the distance measure of one hundred kms. And excluded the correlation between the facilities under the same system along with the ones with patient referrals and exchanges.

4) Creation of facility network graphs using DAGs: Using the correlation data, DAGs were created at an overall facility level and for each service line[18] (the departments in each hospital) using *networkx*. Each facility represents a node and the edges between two nodes are created if the partial correlation is above the threshold. Based on the distance and the edge weights, the connected components are extracted. These connected components can be treated as community of facilities that are related or competitive to each other. The DAG for the Pulmonary service line can be found in Fig. 5.2.

5) Extracting the competitors from each DAG: Once the facility network graphs were created, the neighboring nodes with negative weights were tagged as competitors. The competitors for each facility were extracted.

6) Ranking the competitors: The competitor facilities were ranked based on the patient encounter volume. This resulted in a list of facilities and their competitors with the edge weights from DAG. Table 1 shows the output of sample competitor list.

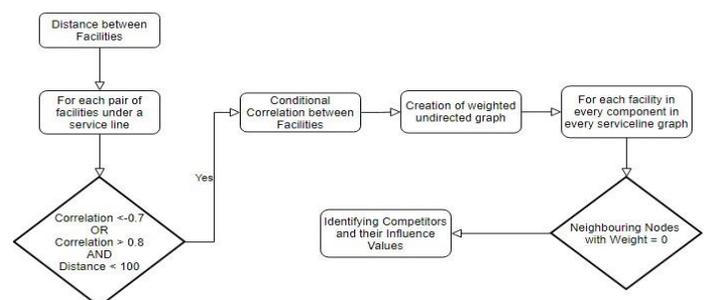

Fig. 5.1 Competitor Identification Process Flow

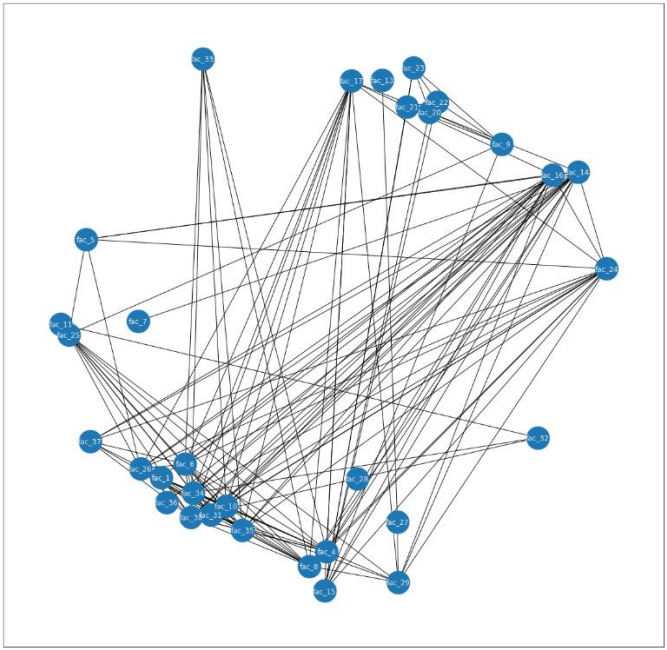

Fig. 5.2 DAG for Pulmonary service line

TABLE I
SAMPLE COMPETITOR LIST

| Service Line | Facility | Month Year | Competitors |
|---|---|---|---|
| Cardiology | fac1 | 2020-02 | [(fac4, 0.08178), (fac8, 0.00119), (fac3, 0.00032), (fac9, 0.00023)] |
| Pulmonary | fac1 | 2020-05 | [(fac7, 0.05445), (fac1, 0.03775)] |
| Cardiology | fac2 | 2021-03 | [(fac6, 0.08178), (fac2, 0.00119), (fac3, 0.00032)] |
| Neurology | fac2 | 2020-02 | [(fac6, 0.08178), (fac8, 0.00119), (fac3, 0.00032), (fac9, 0.0005)] |
| Cardiology | fac3 | 2021-04 | [(fac1, 0.08442), (fac4, 0.00038), (fac9, 0.00013)] |

*B. Machine Learning - Regression Framework*

To get the key attributes of market share it is essential to understand the market. Multiple regression models were tried to learn the trends of market share. The Random Forest regression model turned to be the optimal one for the data used. After extensive hyperparameter tuning, the parameters for random forest resulted in MAPE of 10.1%.

1) Dataset Summary: The training dataset consists of 24 months of data (Jan 2020 – Dec 2021) and the test dataset is of 3 months (Jan 2022 – Mar 2022). This dataset has patient encounters span across 60 different facilities from the Washington state. After feature engineering, the resulting dataset had ~145 features.

2) Loss Function: The range of target variable (market share) is between 0 and 100. As the target is bound in a specific range the widely accepted Mean Square Error (MSE) is considered as the loss function in this experiment.

3) Regression Analysis: Experimented with 3 different regression algorithms viz., Linear Regression (LR), Random Forest Regressor (RF)[19] and XGBoost Regressor (XGB)[20]. Root Mean Square Error (RMSE) and Mean Absolute Percentage Error (MAPE) were considered as the choice of error metrics to quantify performance of various models. With multiple shallow iterations found that Random Forest regressor was the better regressor for the data. The LR is considered as the base line model, whereas RF & XGB models are further fine-tuned to identify optimal parameters

4) Hyperparameter Tuning:

    a) As part of the hyperparameter tuning exercise implemented Random Search technique to identify the better parameters for RF & XGB Models

    b) *HyperOpt*[21], sequential model-based optimization principle uses a tree of parzen estimators approach for optimization of hyperparameter search space, which is a better alternative to Random Search, was used for identifying the better set of parameters for RF & XGB models. Table II has the error metrics for Fine-tuned RF & XGB Models along with baseline LR model

    c) k-Fold Cross Validation: 5-Fold cross validation was implemented to test the stability of the model(s) across all sections of data. The RMSE and MAPE at each fold for all the three regression algorithms are attached in Table III. It can be observed that the model results are consistent across multiple folds

TABLE II
ERROR METRICS OF THE REGRESSION MODELS

| Model | Metric | Train | Test |
|---|---|---|---|
| LR | RMSE | 20.65 | 22.67 |
| LR | MAPE | 16.74% | 17.22% |
| RF | RMSE | 17.84 | 18.24 |
| RF | MAPE | 10.12% | 11.03% |
| XGB | RMSE | 19.37 | 20.08 |
| XGB | MAPE | 11.84% | 12.17% |

TABLE III
K-FOLD (5) CROSS VALIDATION ERROR METRICS

| Model | Metric | Fold 1 | Fold 2 | Fold 3 | Fold 4 | Fold 5 | Mean |
|---|---|---|---|---|---|---|---|
| LR | RMSE | 24.31 | 17.22 | 19.25 | 22.27 | 25.53 | 21.71 |
| LR | MAPE | 16.88% | 14.63% | 14.67% | 15.27% | 16.24% | 15.54% |
| RF | RMSE | 23.59 | 22.38 | 18.88 | 23.51 | 17.81 | 21.23 |
| RF | MAPE | 11.44% | 10.48% | 12.68% | 10.37% | 10.36% | 11.07% |
| XGB | RMSE | 23.26 | 24.84 | 17.10 | 22.31 | 18.27 | 21.56 |
| XGB | MAPE | 13.36% | 10.51% | 9.98% | 10.11% | 9.38% | 10.67% |

5) Feature Importance: Extracted the feature importance scores for RF and XGB. Top fifteen features based on feature importance for RF & XGB along with LR models' top features based on coefficient values are shown in Table IV. The inferences from feature importance scores can be made at global level. Hence to have local interpretability, we used

SHAP. It aided in understanding the key drivers at a local(facility) level.

TABLE IV
TOP 15 FEATURES** AT A MODEL LEVEL

| Rank | Linear Regression | Random Forest | XGBoost |
|---|---|---|---|
| 1 | hospown_Government - Hospital District or Authority | LicensedBedCnt | nursavgrate |
| 2 | htype_Critical Access Hospitals | sa_WA_MT SE WA | sa_WA_MT SE WA |
| 3 | sa_WA-MT SE WA | nursavgrate | sa_PGTSND KING |
| 4 | Is_Covid | zip_rank2 | facility_encoded |
| 5 | hospown_Voluntary non-profit - Private | facility_encoded | PatAvgAge |
| 6 | htype_Childrens | zip_rank1 | LicensedBedCnt |
| 7 | nursavgrate | sa_OR SW WA | sa_OR SW WA |
| 8 | sa_OR SW WA | PatAvgAge | sa_WA-MT INWA |
| 9 | hospown_Voluntary non-profit - Church | pyr_HMOManagedCare | pyr_HMOManagedCare |
| 10 | pyr_CharityCare | pyr_MedicaidManagedCare | pyr_MedicareFeeforService |
| 11 | pyr_PremeraBlueCross | pyr_MedicaidFeeforService | sa_PGTSND SOUTH |
| 12 | pyr_MedicaidFeeforService | age60to70 | Zip_rank4 |
| 13 | pyr_MedicaidManagedCare | referral_cnt | pyr_MedicaidFeeforService |
| 14 | pyr_OtherGovernment | physcare_topbox_perc | Age10to20 |
| 15 | pyr_commercialPrivateIndemPPO | baseclass_perc_ed | Numoffac0to10km |

** Added the key definitions of features in appendix

### C. SHAP Analysis

*SHAPley Values* are used to get the local interpretability of all the features that were used to train the model. The SHAP values are extracted at the Facility – Month level. Fig. 5.4 shows the important features across the dataset.

Aggregated SHAP values at facility level and features were ranked to understand the most influencing factors of market share per facility.

In addition, the features were grouped into categories through SMEs expertise and the understanding of the features. The SHAP values were evaluated at the level of these groups as well. Brought out the top fifteen attributes that affect the market share for each facility. The groups are as follows: Physician Encounters, Payor Groups, Nurse ratings, Encounter Types, Service Lines, Zip Level Encounters, Service Area, Hospital Types, Facility Ratings.

## VI. RESULTS

Two different evaluation frameworks are considered for the two approaches of competitor identification and the regression framework.

Annotation Agreement Factor: The annotation exercise was formulated using the 3 human annotators who are SMEs from the respective Business Units.

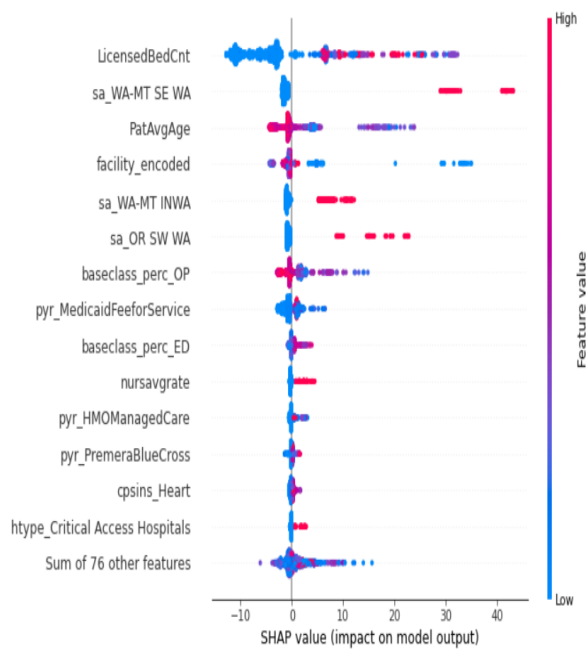

Fig. 5.4 Bee Swarm Plot of overall SHAP values

The scale of 1 to 5 is considered with 1 being the lowest agreement score and 5 being the highest between SME accepted results and model results. Each component is at least annotated by 2 individual annotators and the final score of the agreement between of the connected components is evaluated as average of the scores. The dataset has 60 facilities and connected components are evaluated at overall facility level and service line level. Mean acceptance score was found to be 3.87 out of 5 for overall facility level and 3.75 out of 5 at service line level. The results are in acceptable levels from the SME's feedback. Table V shows the summarized results of annotation exercise.

TABLE V
ANNOTATION AGREEMENT FACTOR** – MEAN ACCEPTANCE SCORE

| Annotator | Overall Facility Level | Service Line Level |
|---|---|---|
| 1 | 3.96 | 3.81 |
| 2 | 3.74 | 3.68 |
| 3 | 3.91 | 3.76 |
| Overall | 3.87 | 3.75 |

** additional explanation is added in appendix

Mean Absolute Percentage Error (MAPE) and the Root Mean Square Error (RMSE) is used as the evaluation metric for the Regression framework. The fine-tuned regression model Random Forest has the MAPE in range of 10.1% - 11%. With RMSE in range of 19-23

SHAP Value Ranking: The key attributes for three of the facilities (chosen arbitrarily from set of 60 Facilities) can be found in Table VI. It can be observed that the set of features impacting Market Share are varying across facilities. From preliminary SME feedback it is observed that the key drivers obtained through the SHAP output are nearer to the expected attributes which drive the market share when compared to features from the aggregated feature importance scores evaluated at the overall regressor level.

TABLE VI
TOP 15 FEATURES** BY SHAP ON RANDOM FOREST FOR 3 FACILITIES

| Rank | Facility 1 | Facility 2 | Facility 3 |
|---|---|---|---|
| 1 | baseclass_perc_OP | PatAvgAge | sa_WA-MT SE WA |
| 2 | baseclass_perc_ED | baseclass_perc_OP | PatAvgAge |
| 3 | nursecnt | baseclass_perc_ED | baseclass_perc_OP |
| 4 | cpsins_Heart | pyr_SelfPay | zip_rank1 |
| 5 | pyr_MedicareFeeforService | TotalPhysicians | pyr_MedicaidFeeforService |
| 6 | TotalPhysicians | physician_rank3 | baseclass_perc_ED |
| 7 | cpsins_DigestiveHealth | physician_rank2 | cpsins_Heart |
| 8 | age30to40 | pyr_DepartmentofDefense | pyr_WorkerCompensation |
| 9 | cpsins_Cancer | cpsins_Orthopedics | pyr_KaiserPermanente |
| 10 | pyr_Regence | LOSAvg | TotPhysWithAtleast30Encs |
| 11 | age60to70 | pyr_Regence | age20to30 |
| 12 | age40to50 | age80plus | baseclass_perc_IP |
| 13 | physician_rank3 | physician_rank1 | zip_rank5 |
| 14 | LOSAvg | Physician_rank4 | age30to40 |
| 15 | physician_rank1 | Referral_cnt | zip_rank4 |

** Added the key definitions of features in appendix

## VII. CONCLUSION

This paper proposes a two-pronged approach, to identify the key competitor pool of facilities and then understand the key factors driving the market share. The paper tackles typical challenges in competitor detection and proposed data driven framework to identify competitors. The solution relies on two different Machine Learning disciplines, Graph Learning and Regression analysis. The framework developed has been evaluated on multiple US states for its effectiveness and scalability. For future work, it is intended to expand into development of a healthcare strategic simulator, which would empower the users to simulate scenarios of changes in several factors and understand the impacts on encounter and market share of theirs and competitors. This would be achieved by stitching together all the pieces of the puzzles involved like competitor detection, key drivers to market share, future forecasting, evolving market dynamics, etc. The goal of this work is to empower the strategists with all the necessary information, so the hospitals can arrive at effective strategies for a better health of patients.

APPENDIX

TABLE VII
DATA DICTIONARY OF FEATURES USED

| slno | Fields Names | Field Description |
|---|---|---|
| 1 | facility | Facility Name |
| 2 | monyr | Month-Year |
| 3 | facenccnt | Facility, Month-Year level Encounter count |
| 4 | compenccnt | Competitors, Month-Year level Encounter count for each facility in "facility" field |
| 5 | competitors_list | Competitors list for facility in "facility" field |
| 6 | market_share | ratio of facility encounters to total encounters |
| 7 | nursecare (topbox_perc, botbox_perc, intensecare_perc, NICUcare_perc, labourcare_perc) | Nurse - In-Patient Reviews Top Box and Bottom Box |
| 8 | physcare (topbox_perc, botbox_perc, intensecare_perc, NICUcare_perc, labourcare_perc) | Physician - In-Patient Reviews Top Box and Bottom Box |
| 9 | nursavgrate | Average In-Patient Nurse satisfaction rate |
| 10 | phyavgrate | Average In-Patient Physician satisfaction rate |
| 11 | Nurse (NEGperc & POSperc) | Patient Reviews - Nurse - Sentiment |
| 12 | Phys (NEGperc & POSperc) | Patient Reviews - Physician - Sentiment |
| 13 | cpsins (Cancer, DigestiveHealth, Heart, Neuroscience, Orthopedics, Unmapped, WomenandChildren, exiscnt, AllOther) | Facility, Month-Year, CPS INSTITUTES level Encounter counts |
| 14 | zip (rank1 to rank10) | Zipcodes Ranked by encounter count for Facility, Month-Year combination |
| 15 | DRG (rank1 to rank10) | Top 10 DRG Codes Ranked by encounter count for Facility, Month-Year combination |
| 16 | physicians (rank1 to rank10) | Top 10 Physicians Ranked by encounter count for Facility, Month-Year combination |
| 17 | TotalPhysicians | Total Physicians attending at Facility, Month-Year combination |
| 18 | TotalPhysiciansWithAtleast30Encs | Total Physicians attending at Facility, Month-Year combination with atleast 30 Encounters |
| 19 | pyr (commercialPrivateIndemPPO, DepartmentofDefense, DeptofVeteransAffairs, HMOManagedCare, HealthExchange, IndianHealthServiceofTribe, KaiserPermanente, MedicaidFeeforService, MedicaidManagedCare, MedicareFeeforService, MedicareManagedCare, CharityCare, OtherGovernment, PremeraBlueCross, Regence, SelfPay, WorkerCompensation) | Facility, Month-Year, Payer Group level Encounter counts |
| 20 | popgrow (0to10, 10to20, 20to30, 30to40, 40to50, 50to60, 60to70,70to80, 80Plus) | Historical Population Growth by Age Group(Facility, Year level) |
| 21 | 5yrprojgrow (0to10, 10to20, 20to30, 30to40, 40to50, 50to60, 60to70,70to80, 80Plus) | Population Growth 5 yr Projection by Age Group(Facility serving zipcodes, Year level) |
| 22 | HospitalType | Hospital Attributes |
| 23 | HospitalOwnership | Hospital Attributes |
| 24 | EmergencyServices | Hospital Attributes |
| 25 | numoffac0to10km | Number of facilities within 10KM Radius |
| 26 | baseclass (Emergency Department, Inpatient, Outpatient) | Encounter count at Facility, Month-Year, Base Class Desc level |
| 27 | age (0to10, 10to20, 20to30, 30to40, 40to50, 50to60, 60to70,70to80, 80Plus) | Encounter count at Facility, Month-Year, Age Groups level |
| 28 | referral (count, percentages in a month) | Referrals |
| 29 | LOS (Avg, StdDev, max) | Length of stay |
| 30 | RACE (AmericanIndianORAlaskanNative, Asian, BlackORAfricanAmerican, 2 or more races, NativeHawaiianOrPacificIslander, Unknown, White) | Encounter count at Facility, Month-Year, Race level |
| 31 | Is_Covid | Flag to describe Covid Support facilities |

Annotation Agreement Factor aggregations: In the current exercise considered the creation of connected components from DAGs at 2 different categories i.e., at Overall facility level & Service line for which SMEs annotation was performed. Each category has a varied number of connected components which are ranked by annotators with a prior that each component is scored by at-least two annotators. Table IX just illustrates how the raw scores per component for each category are recorded by the annotators and then aggregated at multiple levels

TABLE IX
ANNOTATION AGREEMENT FACTOR – RAW DATA ILLUSTRATION PER CATEGORY

| Component | Annotator_1 | Annotator_2 | Annotator_3 | Mean |
|---|---|---|---|---|
| 1 | 4 | 3 | - | 3.5 |
| 2 | 3 | 2 | 3 | 2.67 |
| 3 | - | 5 | 4 | 4.5 |
| 4 | 4 | - | 4 | 4 |
| 5 | 5 | - | 3 | 4 |
| 6 | 3 | 2 | - | 2.5 |
| ⋮ | ⋮ | ⋮ | ⋮ | ⋮ |
| N | 2 | 1 | 2 | 1.67 |
| Mean Ratings | x.yy | y.zz | z.aa | t.pp |